\documentclass[10pt,twocolumn,letterpaper]{article}

\usepackage{cvpr}      
\definecolor{cvprblue}{rgb}{0.21,0.49,0.74}
\usepackage[pagebackref,breaklinks,colorlinks,allcolors=cvprblue]{hyperref}

\usepackage{makecell}

\usepackage{xcolor}
\usepackage{colortbl}   
\usepackage[table,xcdraw]{xcolor}  
\usepackage{multirow}
\usepackage{booktabs}
\usepackage{array}
\def\paperID{8799} 
\def\confName{CVPR}
\def\confYear{2026}

\title{Frequency-Modulated Visual Restoration for Matryoshka Large Multimodal Models}
\author{
Qingtao Pan$^{1}$ \quad
Zhihao Dou$^{1}$ \quad
Shuo Li$^{1}$\thanks{Corresponding author: shuo.li11@case.edu} \\
\\
$^{1}$Case Western Reserve University
}

\begin{document}
\maketitle
\begin{abstract}
Large Multimodal Models (LMMs) struggle to adapt varying computational budgets due to numerous visual tokens. Previous methods attempted to reduce the number of visual tokens before or within LLMs. However, these strategies inevitably result in the loss of visual semantic. To address these issues, we introduce \textbf{FMVR}, a plug-and-play and extremely simple \textbf{F}requency-\textbf{M}odulated \textbf{V}isual \textbf{R}estoration strategy to boost the reasoning ability of LMMs under visual token reduction. Specifically, FMVR disentangles the visual representation of fewer visual tokens into low- and high-frequency components through AvgPool and MaxPool. The derived frequencies are subsequently modulated using lightweight learnable parameters. The high-frequency from AvgPool acts as a saliency filter to enhance saliency visual semantics, while the low-frequency from MaxPool acts as an anti-saliency filter to strengthen weak visual semantics. It enables the preservation of visual semantics dominated by few visual tokens and the restoration of diluted visual semantics. Additionally, we inject FMVR into Matryoshka Representation Learning to learn coarse-to-fine visual token sets, thus enabling to elastically adjust the number of visual tokens during inference while maintaining comparable performance. Experiments across 10 image-based and 4 video-based bench marks demonstrate that FMVR-LLaVA reduce the FLOPs of LLaVA-1.5-7B by 89\%, while maintaining almost 100\% of the original accuracy. The code will be open.
\end{abstract}    
\section{Introduction}
\label{sec:intro}

Empowered by the linguistic reasoning capacity of large language models (LLMs) \cite{refi1,refi2,refi3,refi4}, Large Multimodal Models (LMMs) \cite{refi5,refi6,refi7,refi8} extend the reasoning capabilities of LLMs to the visual domain, achieving joint understanding across visual and linguistic modalities. These LMMs typically utilize a visual encoder to transform images into discrete visual tokens, which are subsequently fused with textual tokens and processed by a large language model backbone.

\begin{figure}[t]
\centering
\includegraphics[width=\linewidth]{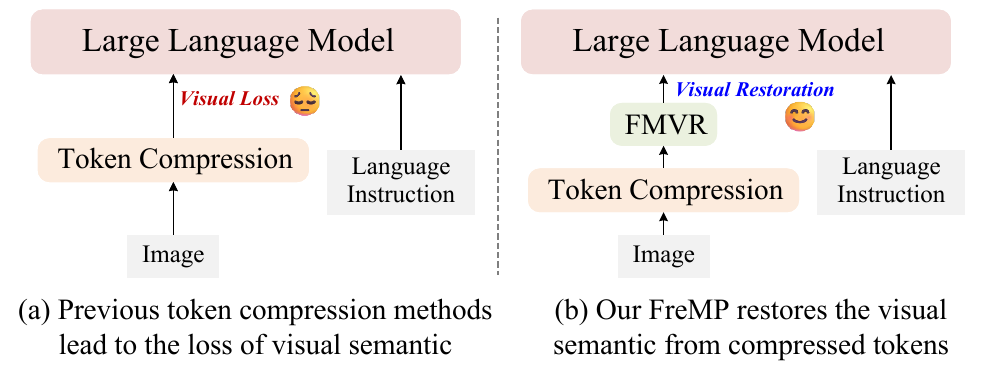}
\caption{Our FMVR (b) can restore the visual semantics from compressed tokens, alleviating the loss of visual contents in previous token compression methods (a).}
\label{fig1}
\end{figure}

In practice, the token budget available to different applications can vary considerably depending on computational constraints and real-time performance requirements. In resource-limited scenarios, a smaller token capacity is often mandated to maintain processing efficiency. Nevertheless, most LMMs employ relatively high visual token count per image. Given that both computational time and memory consumption increase quadratically with the input sequence length \cite{refi9,refi10,refi11,refi12,h1}, using a large number of visual tokens can significantly exacerbate the computational overhead of LMMs. Moreover, in video \cite{refi13,refi14,refi15} or high-resolution scenarios \cite{refi16,refi17,refi18}, the number of required visual tokens can increase dramatically. Consequently, excessive tokenization not only slows inference but also, due to the lack of adaptive visual token allocation at deployment time, restricts the scalability and flexibility of LMMs in practical applications \cite{refi19}.

To alleviate this issue, prior studies \cite{refi20,refi21,refi22,refi23} have turned to visual token compression, aiming to discard redundant visual tokens while preserving essential semantic information. Despite their effectiveness, these methods inherently yield outputs with a fixed sequence length, without flexibility to adjust the number of visual tokens. Such rigidity hinders the ability to dynamically balance the trade-off between visual tokens and computational cost, which is particularly crucial for computational limitations. Inspired by Matryoshka Representation Learning \cite{refi24}, several studies \cite{refi25,refi26} have investigated training a unified model capable of processing variable numbers of visual tokens (Fig. \ref{fig1}(a)), enabling it to adapt dynamically to different input granularities and computational budgets. However, these methods lead to semantic loss for detailed images, which hampers the model’s ability to capture subtle visual content, leading to poor image understanding performance.

\begin{figure}[t]
\centering
\includegraphics[width=\linewidth]{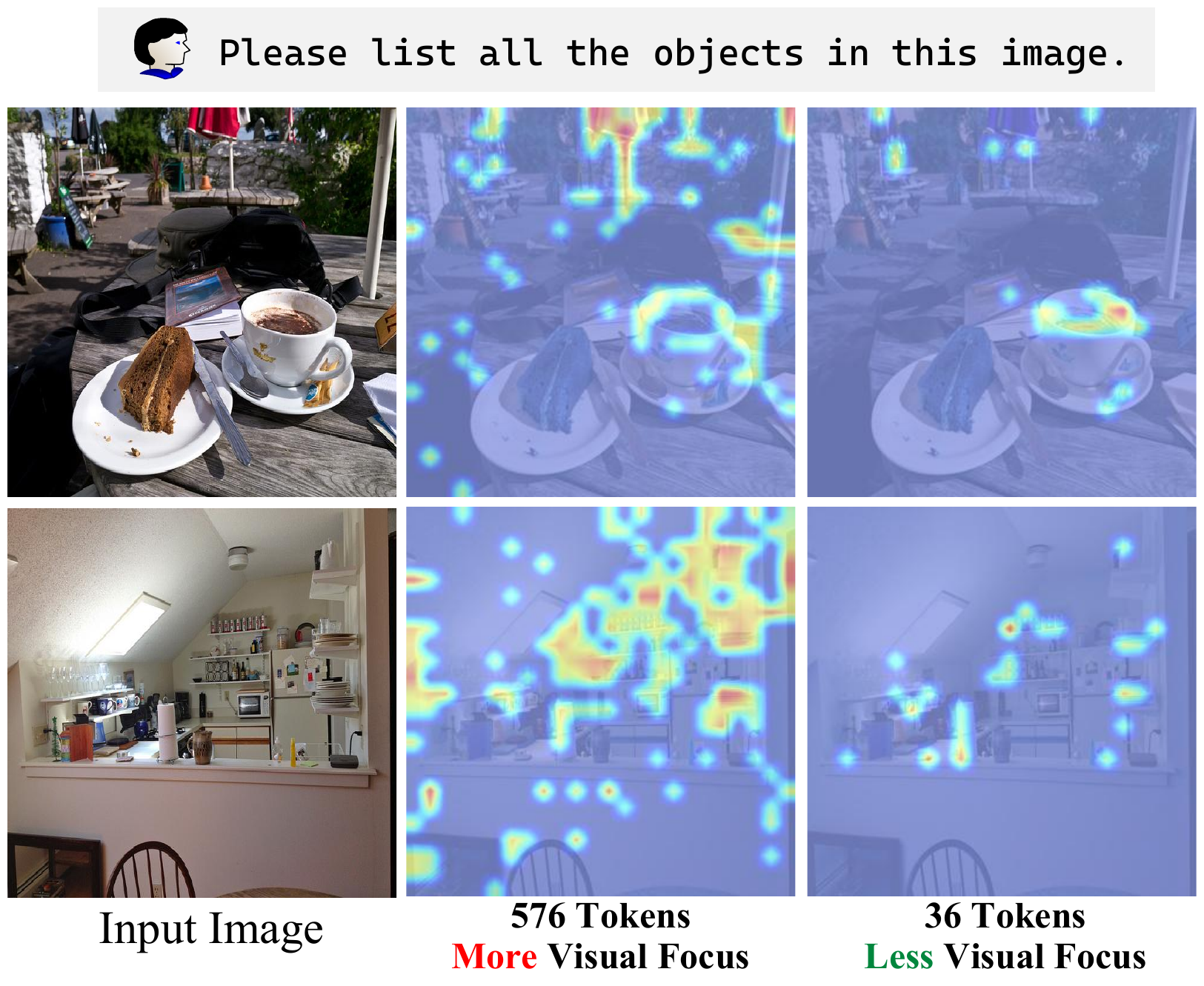}
\caption{Grad-CAM visualization (576 and 36 visual tokens) shows that the reduction of visual tokens leads to a noticeable degradation in visual focus.}
\label{fig2}
\end{figure}

To further substantiate this, we performed a preliminary experiment analyzing the impact of reduced visual tokens on imaging understanding. As shown in Fig. \ref{fig2}, the Grad-CAM reveals a noticeable degradation in visual contents when the number of visual tokens is reduced (576→36). In contrast, a full set of visual tokens (576) exhibits dense visual attention, focusing on nuanced visual regions. This observation highlights the importance of preserving detailed visual contents under token reduction.

Motivated by the above observation, we seek to tackle a pivotal issue: \textit{How to enhance LMM's reasoning capability under limited visual token budgets.} To achieve this, inspired by the frequency components in feature restoration \cite{refi27}, we propose FMVR, a \textbf{F}requency-\textbf{M}odulated \textbf{V}isual \textbf{R}estoration strategy that recovers visual semantics from compressed visual tokens. Concretely, FMVR can disentangle the visual representation into distinct low- and high-frequency components through AvgPool and MaxPool. These frequency components are then refined through lightweight and learnable modulation parameters, The high-frequency from AvgPool acts as a saliency filter to enhance saliency visual semantics, while the low-frequency from MaxPool act as an anti-saliency filter to strengthen weak visual semantics. Subsequently, we inject FMVR into Matryoshka Representation Learning (MRL) to construct nested visual tokens for LMM training, enabling LMM to elastically adjust visual tokens according to computational constraints while retaining effective reasoning ability. We summarize our contributions as follows:
\begin{itemize}
    \item We conduct an analysis of the phenomenon where visual token reduction impairs the model’s ability to focus nuanced visual semantics.
    \item The idea of visual restoration in token compressed LMM is first formulated to alleviate visual semantic degradation induced by token sparsity.
    \item FMVR, a plug-and-play and extremely simple visual token semantic restoration method, is proposed to highlight diluted visual semantics by disentangling and modulating the frequency components derived from limited visual tokens, thus enriching visual representations.
    \item Combining FMVR and Matryoshka Representation Learning, it achieves elastically and efficient visual reasoning across varying computational budgets.
    \item FMVR-LLaVA consistently outperforms prior state-of-the-art methods in various image and video understanding benchmarks, while exploring the trade-off between the accuracy and the number of visual tokens.
\end{itemize}

\section{Related work}
\label{sec:formatting}

\begin{figure*}[t]
\centering
\includegraphics[width=\linewidth]{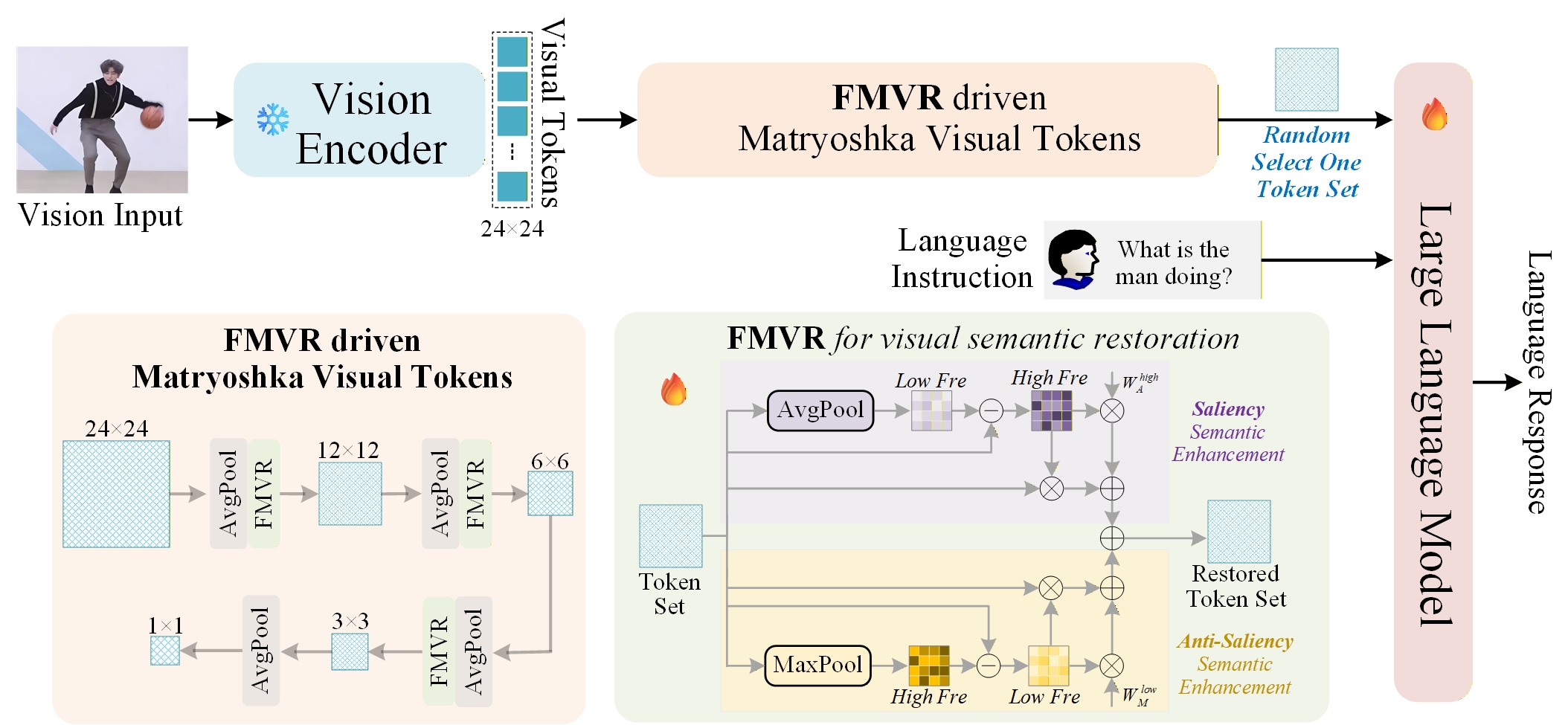}
\caption{\textbf{Illustration of FMVR-LLaVA.} The FMVR is injected into MRL to construct nested visual tokens, where FMVR is used to enhance the visual semantics of each visual token set, thus forming a set of reinforced nested visual tokens for LLM training.}
\label{fig3}
\end{figure*}

\textbf{Large Multimodal Models.} The advancements in Large Language Models (LLMs) \cite{refi1,refr1,refi2,refi3,refi4,h2} have catalyzed a paradigm shift in artificial intelligence, inspiring researchers to extend linguistic reasoning to the multimodal domain, including image understanding \cite{h3,h4,h5} and medical imaging \cite{h6,h7}. In such models, visual inputs are discretized into token representations to fully leverage the inference capabilities of LLMs. However, visual token representations are dense and spatially redundant, resulting in a significant token expansion. For instance, LLaVA-1.5 \cite{refi5} transforms a 336×336 image into 576 tokens. Therefore, improving inference efficiency has emerged as a pivotal factor for the scalability and practical deployment of LMMs.

\noindent
\textbf{Visual Token Reduction.} To achieve compact visual representations, a variety of approaches have been proposed to reduce the number of visual tokens while retaining essential semantic information. Early studies \cite{refr6,refr7,refr8,refr9,refr10} primarily relied on the Q-Former \cite{refr8}, which transforms visual embeddings into a set of fixed-length learnable queries. Although these approaches help alleviate performance degradation induced by token compression, they remain constrained by a prespecified pruning threshold. Recent methods \cite{refi25,refi26}, inspired by Matryoshka Representation Learning \cite{refi24}, introduce a hierarchical nesting mechanism for visual tokens, enabling models to represent visual information at multiple granularities. However, these methods inevitably cause information degradation, thereby impairing the model’s capacity to preserve fine-grained visual semantics when operating with a limited number of visual tokens. The proposed FMVR can alleviate visual information degradation caused by token reduction by decomposing visual representations into low- and high-frequency components, enabling the visual semantic restoration.
\section{Method}
The FMVR-LLaVA, as illustrated in Fig. \ref{fig3}, proposes a novel visual semantics restoration mechanism, FMVR, to recover nuanced visual contents under compressed visual tokens. By combining the idea of Matryoshka Representation Learning \cite{refi24}, FMVR-LLaVA can support elastically visual token scaling in inference, achieving a favorable balance between computational efficiency and reasoning performance. To be specific, instead of leveraging raw compressed visual tokens, FMVR decomposes the compressed visual representations into low- and high-frequency components to highlight diluted visual semantics.

\subsection{Preliminaries}
\textbf{Problem Formulation.} Typically, LMMs comprise three core components: vision encoder $f_{V}$, modality projector $f_{M}$, and LLM $f_{LLM}$. For simplicity, we define that visual features $H^V=f_{M}(f_{V}(X^{V}))\in\mathbb{R}^{M\times d}$, where $X^{V}$ is the input image, $M$ denotes the length of vision tokens, and $d$ is the hidden dimension size; textual features $H^{T}\in{R}^{N\times d}$ where $N$ denotes length of text tokens. In practice, $M$ is generally much larger than $N$. The LLM utilizes the visual and textual features to generate output responses $X^{R}$. However, the excessive number of visual tokens significantly increases computational cost, resulting in inefficient inference \cite{refi21}. Hence, vision token compression can effectively reduces $M$ could ameliorate this inefficiency.

\noindent
\textbf{Matryoshka Representation Learning (MRL).} MRL is designed to jointly optimize embeddings across multiple dimensional scales. Let $E$ denote the set of target embedding dimensions. MRL introduces a learnable model for each subset $v_{1:e}\in\mathbb{R}^e$, where $e\in E$. This strategy allows each truncated representation to be independently optimized under the shared objective. The formal objective function of MRL is defined as:
\begin{equation}
\mathcal{L}_{\mathrm{MRL}}=\sum_{e\in E}c_e\mathcal{L}\left(\boldsymbol{W}^{(e)}\cdot f(x;\theta_f)_{1:e};y\right),
\end{equation}
where $\mathcal{L}$ refers to the objective loss and $y$ the ground-truth label. During optimization, MRL simultaneously performs forward and backward updates across all $E$ submodels. Once trained, it enables inference with any embedding size $d\leq E$, achieving scalable deployment depending on resource constraints. Building on this concept, we aim to extend it to LMMs that can dynamically adjust visual tokens for elastic multimodal reasoning.

\subsection{FMVR driven Matryoshka Visual Tokens}
Matryoshka-style nested visual tokens is constructed by progressively performing 2×2 pooling with a stride of 2. To be specific, the CLIP visual encoder \cite{refm1} processes an input image $X^{V}$ as 24×24 visual tokens. Such 24×24 visual tokens are gradually pooled into 12×12, 6×6, 3×3, and 1×1 visual tokens, thus constructing $S$ nested visual tokens (i.e., $S_{i}\in\{1,9,36,144,576\}$), where fewer tokens are obtained by progressively compressing denser tokens. To enhance the visual semantics of each visual token set, we propose FMVR and inject it into each 2×2 pooling operation.

\subsection{FMVR for visual semantic restoration}
Visual token compression in LMMs tends to the loss of visual information, which is essential for visual understanding. To mitigate this, we introduce FMVR, which highlights subtle visual semantics by disentangling and modulating the frequency components derived reduced visual tokens. As shown in Fig. \ref{fig3}, FMVR contains two distinct pooling units (AvgPool or MaxPool). The AvgPool unit extracts high-frequency to emphasize salient visual semantics. The MaxPool unit acts as an anti-saliency filter to recover weak visual semantics, avoiding strong salient semantics dominating.

\noindent
\textbf{AvgPool Unit for Saliency Semantics Enhancement.} To mitigate semantic dilution caused by 2×2 pooling operation in matryoshka visual token construction, we apply AvgPool \cite{refm2} to extract the low-frequency component $X^l_A$ (captures the global semantic) of each matryoshka visual token set $X\in\mathbb{R}^{C\times H\times W}$. Then we compute the residual between $X$ and $X^l_A$ to obtain the high-frequency component $X^h_A$, which acts as a saliency filter to highlight saliency visual semantics, calculate as follows:
\begin{equation}
X^h_A=X-X^l_A,\quad X^l_A=\mathrm{AP}(X)
\end{equation}
where $\mathrm{AP}(\cdot)$ is AvgPool operation. The high-frequency component $X^h_A$ is then modulated by learnable parameters,
\begin{equation}
\hat{X}^h_A=W_A^h\cdot X^h_A,
\end{equation}
where $W_A^h$ denotes learnable channel-wise parameters that are optimized by backpropagation to select useful frequency subbands. Additionally, the high-frequency is repurposed as attention maps, which are applied multiplicatively to the original input features to boost the activation of visually significant areas. Finally, the high-frequency component and attention weighted features are summed to produce the final representation:
\begin{equation}
\hat{X}_A=\hat{X}^h_A+X^h_A\cdot X
\end{equation}
However, when a strongly salient semantic dominates the visual representation, weaker semantics are likely to remain overshadowed.

\noindent
\textbf{MaxPool Unit for Anti-saliency Semantics Enhancement.} To address the dominance of over-salient semantics, the MaxPool is utilized for weaker visual semantics (i.e., anti-saliency Semantics) enhancement. Specifically, we first extract high-frequency component $X^h_M$ that represents relatively significant semantics through MaxPool. Then we compute the residual between $X$ and $X^h_M$ to capture the differences between non-salient and salient semantics. Intuitively, this residual denotes the low-frequency $X^l_M$, which acts as an anti-saliency filter that suppresses the dominant sementics and explicitly enhances the latent weak semantics.
\begin{equation}
X^l_M=X-X^h_M,\quad X^h_M=\mathrm{MP}(X),
\end{equation}
where $\mathrm{MP}(\cdot)$ is MaxPool operation. Similarity, the low-frequency $X^l_M$ is also modulated using learnable parameters:
\begin{equation}
\hat{X}^l_M=W_M^l\cdot X^l_M,
\end{equation}
where $W_M^l$ is learnable modulation parameters. The low-frequency is employed as attention maps to reactivate suppressed semantics.The output of MaxPool Unit is,
\begin{equation}
\hat{X}_M=\hat{X}^l_M+X^l_M\cdot X
\end{equation}
The final restored token set is the sum of $\hat{X}_A$ and $\hat{X}_M$.

\section{Experiments}
\subsection{Experimental Setup}

\begin{table*}[t]
\centering
\caption{Performance comparison on LLaVA-1.5-7B across 10 image-based benchmarks.  ‘\#VisionTokens’ is the number of vision tokens. $\Delta$ rows show the change vs.\ LLaVA baseline.}
\renewcommand{\arraystretch}{1.1}
\setlength{\tabcolsep}{3pt} 
\small
\resizebox{0.95\textwidth}{!}{
\begin{tabular}{l|c| c c c c c c c c c c | c}
\toprule
\textbf{Methods} &
\makecell[c]{\textbf{\#Vision}\\\textbf{Tokens}} &
\textbf{VQAv2} & \textbf{GQA} & \textbf{VisWiz} & \textbf{SQA\textsuperscript{IMG}} &
\textbf{VQA\textsuperscript{Text}} & \textbf{POPE} & \textbf{MME} & \textbf{MMB\textsuperscript{EN}} & \textbf{MMB\textsuperscript{CN}} & \textbf{MMVet} &
\makecell[c]{\textbf{Avg.}\\\textbf{(\%)}} \\
\midrule
BLIP-2~\cite{refe11}        & 32  & 65.0 & 41.0 & 19.6 & 61.0 & 42.5 & 85.3 & 1293.8 & 31.2 & -- & 22.4   & --   \\
InstructBLIP~\cite{refe12} & 32  & 72.4 & 49.2 & 34.5 & 60.5 & 50.1 & 86.1 & 1391.4 & 36.0 & 23.7 & 26.2 & 50.8 \\
IDEFICS-9B~\cite{refe13}        & 64  & 50.9 & 38.4 & 35.5 & 53.5 & 25.9 & 81.9 & 1177.3 & 48.2 & -- & 30.0 & --   \\
IDEFICS-80B~\cite{refe13}       & 64  & 60.0 & 45.2 & 36.0 & 61.8 & 30.9 & 66.0 & 1518.2 & 54.5 & -- & 39.7   & --   \\
Qwen-VL~\cite{refe14}            & 256 & 78.8 & 59.3 & 35.2 & 67.1 & 63.8 &  70.0 & 1487.6 & 38.2 & 7.4 & 13.0 & 50.7 \\
Qwen-VL-Chat~\cite{refe14}   & 256 & 78.2 & 57.5 & 38.9 & 68.2 & 60.7 & 74.9 & 1487.5 & 60.6 & -- & 47.3 & -- \\
SPHINX~\cite{refe15}             & 289 & 78.1 & 62.6 & 39.9 & 69.3 & 51.6 & 80.7 & 1476.1 & 66.9 & -- & 36.0   & --   \\
mPLUG-OwL2~\cite{refe16}      & 1024 & 79.4 & 56.1 & 54.5 & 68.7 & 54.3 & 84.6 & 1450.2 & 64.5 & -- & 36.2 & --   \\
LLaVA-v1.5~\cite{refi5}   & 576 & 78.5 & 62.0 & 50.0 & 66.8 & 58.2 & 85.9 & 1510.7 & 64.3 & 58.3 & 30.5 & 63.0 \\
\midrule
\multicolumn{13}{l}{\textcolor{cyan!60!black}{\textbf{LLMs with vision token reduction methods}}}\\
FastV~\cite{refe17}         &  192  & 67.1 & 52.7   & 46.7   & 65.2 & 52.5 & 64.8 & 1312.4 & 61.2   & 55.4   & 27.7   &  55.9  \\
PyramidDrop~\cite{refe18}       &  192 & 76.1 & 57.1   & 51.3   & 70.2 & 56.1 & 82.3 & 1462.3 & 63.2   & 56.3   & 30.5   & 61.6   \\
SparseVLM~\cite{refe19}        &  192 & 75.6 & 57.6 & 51.6 & 67.2 & 56.1 & 83.6 & 1382.8 & 62.5 & 56.7 & 31.5 & 61.2 \\
Prumerge+~\cite{refi22}              & 144 & 76.8 & 59.3 & 49.8 & 68.3 & 57.1 & 84.0 & 1462.4 & 64.9 & 53.2 & 30.1 & 61.7 \\
TRIM~\cite{refe20}       & 128 & 75.4 & 58.4 & 51.6 & 68.6 & 52.2 & 85.3 & 1413.4 & 63.0 & 52.3 & 29.9 & 60.7 \\
VisionZip~\cite{refe21}              & 192 & 76.8 & 59.3 & 51.2 & 68.9 & 57.3 & 85.3 & 1469.3 & 63.0 & 57.0 & 31.7 & 62.4 \\
DART~\cite{refe22}       & 128 & 74.7 & 57.9 & 52.8 & 69.1 & 56.3 & 80.4 & 1408.7 & 60.7 & 57.3 & 30.9 & 61.1 \\
DivPrune~\cite{refe23}       & 128 & 76.0 & 59.4 & 52.8 & 68.6 & 55.9 & 87.0 & 1405.1 & 61.5 & 54.8 & 30.6 & 61.7 \\
CDPruner~\cite{refe24}       & 128 & 76.6 & 59.9 & 52.8 & 69.0 & 56.2 & 87.7 & 1431.4 & 63.1 & 55.0 & 32.8 & 62.5 \\
VisPruner~\cite{refe25}       & 128 & 75.8 & 58.2 & 52.7 & 69.1 & 57.0 & 84.6 & 1461.4 & 62.7 & 57.3 & 33.7 & 62.4 \\
MQT-LLaVA~\cite{refi26}       & 144 & 76.4 & 61.4 & 52.0 & 67.5 & 53.4 & 83.9 & 1446.4 & 64.4 & 57.2 & 29.9 & 61.8 \\
M3~\cite{refi25}        & 144 & 78.5 & 61.3 & 53.1 & 65.2 & 54.8 & 87.0 & 1451.5 & 66.4 & 56.3 & 30.2 & 62.5 \\
\midrule
\multicolumn{13}{l}{\textcolor{cyan!60!black}{\textbf{Ours}}}\\
 & 1 & 68.3 & 55.2 & 49.7 & 68.6 & 49.2 & 81.1 & 1284.8 & 60.7 & 53.4 & 26.4 & 57.7 \\
 & 9 & 74.5 & 59.1 & 50.7 & 69.9 & 50.8 & 84.1 & 1415.0 & 64.2 & 57.5 & 29.0 & 61.1 \\
\textbf{FMVR-LLaVA} & 36 & 76.5 & 60.9 & 52.9 & 69.5 & 55.3 & 85.9 & 1452.5 & 65.2 & 58.3 & 32.2 & 62.9 \\
 & 144 & 78.6 & 62.3 & 55.1 & 69.7 & 55.5 & 86.4 & 1473.9 & 65.8 & 57.6 & 33.4 & \textbf{63.8} \\
 & 576 & 79.2 & 63.0 & 56.5 & 68.9 & 57.8 & 87.5 & 1510.1 & 65.9 & 58.0 & 34.3 & \textbf{64.7} \\
\quad $\Delta$ vs.\ LLaVA-v1.5 & 144 & {\color{red}+0.1} & {\color{red}+0.3} & {\color{red}+5.1} & {\color{red}+2.9} & {\color{blue}-2.7} & {\color{red}+0.5} & {\color{blue}-36.8} & {\color{red}+1.5} & {\color{blue}-0.7} & {\color{red}+2.9} & {\color{red}+0.8} \\
\bottomrule
\end{tabular}}
\label{table1}
\end{table*}

\begin{table*}[t]
\centering
\caption{Ablation study of applying different pooling unit in FMVR.}
\small
\setlength{\tabcolsep}{4pt} 
\renewcommand{\arraystretch}{1.1} 
\small
\resizebox{0.85\textwidth}{!}{
\begin{tabular}{l| c c c c c c c c c c | c}
\toprule
\textbf{Method} & \textbf{VQAv2} & \textbf{GQA} & \textbf{VisWiz} & \textbf{SQA\textsuperscript{IMG}} &
\textbf{VQA\textsuperscript{Text}} & \textbf{POPE} & \textbf{MME} & \textbf{MMB\textsuperscript{EN}} & \textbf{MMB\textsuperscript{CN}} & \textbf{MMVet} & \textbf{Avg.(\%)} \\
\midrule
M3 (Baseline) & 78.5 & 61.3 & 53.1 & 65.2 & 54.8 & 87.0 & 1451.5 & 66.4 & 56.3 & 30.2 & 62.5 \\
\midrule
\rowcolor{gray!15} FMVR (Ours) & 78.6 & 62.3 & 55.1 & 69.7 & 55.5 & 86.4 & 1473.9 & 65.8 & 57.6 & 33.4 & \textbf{63.8} \\
FMVR w/o AvgPool & 77.9 & 62.0 & 54.6 & 68.1 & 54.3 & 85.7 & 1459.6 & 65.4 & 56.7 & 32.5 & 63.0 \\
FMVR w/o MaxPool & 78.3 & 62.6 & 53.3 & 68.4 & 55.0 & 86.9 & 1463.7 & 65.8 & 57.1 & 32.6 & 63.3 \\
\bottomrule
\end{tabular}
\label{table2}}
\end{table*}

\begin{table}[t]
\centering
\caption{Efficiency analysis of different numbers of vision tokens on LLaVA-1.5-7B.}
\small
\setlength{\tabcolsep}{4pt} 
\renewcommand{\arraystretch}{2.0} 
\footnotesize
\begin{tabular}{l| c  c  c  c | c}
\toprule
\textbf{Methods} & \textbf{Tokens} & \makecell[c]{\textbf{FLOPs}\\\textbf{(TB)}} & \makecell[c]{\textbf{Prefill}\\\textbf{Time (ms)}} & \makecell[c]{\textbf{Total}\\\textbf{Memory (GB)}} & \makecell[c]{\textbf{Avg}\\\textbf{(\%)}} \\
\midrule
LLaVA-v1.5 & 576 & 8.0 &  58.1 & 21.6 & 63.0 \\
\midrule
\multirow{4}{*}{FMVR}
 & 144 & \makecell[c]{2.2\\\color{red}(×3.6)} & \makecell[c]{19.5\\\color{red}(×3.0)} & \makecell[c]{15.0\\\color{red}(×1.4)} & \textbf{63.8} \\

 & 36 & \makecell[c]{0.9\\\color{red}(×8.9)} & \makecell[c]{18.0\\\color{red}(×3.2)} & \makecell[c]{13.8\\\color{red}(×1.6)} & 62.9 \\

 & 9 & \makecell[c]{0.5\\\color{red}(×16)} & \makecell[c]{17.7\\\color{red}(×3.3)} & \makecell[c]{13.6\\\color{red}(×1.6)} & 61.1 \\

 & 1 & \makecell[c]{0.4\\\color{red}(×20)} & \makecell[c]{17.6\\\color{red}(×3.3)} & \makecell[c]{13.5\\\color{red}(×1.6)} & 57.7 \\
\bottomrule
\end{tabular}
\label{table3}
\end{table}

\begin{table}[t]
\centering
\caption{FLOPs of each component in FMVR-LLaVA.}
\footnotesize
\setlength{\tabcolsep}{5pt}
\renewcommand{\arraystretch}{1.2}
\begin{tabular}{l|cccc|c}
\toprule
\textbf{Methods} & \makecell[c]{\textbf{Vision}\\\textbf{Encoder}} & \textbf{Projection} & \textbf{FMVR} & \textbf{LLM} & \textbf{Total} \\
\midrule
LLaVA-v1.5  & 0.349 & 0.024 & - & 8.0 & 8.37 \\
FMVR\color{blue}\textsuperscript{576} & 0.349 & 0.024 & 6.4e-5 & 8.0 & 8.37 \\
\bottomrule
\end{tabular}
\label{table4}
\end{table}

\begin{table}[t]
\centering
\caption{Efficiency analysis of different pruning methods on LLaVA-1.5-7B.}
\small
\setlength{\tabcolsep}{4pt} 
\renewcommand{\arraystretch}{1.1} 
\footnotesize
\begin{tabular}{l| c  c  c | c}
\toprule
\textbf{Methods} & \textbf{Tokens} & \makecell[c]{\textbf{FLOPs}\\\textbf{(TB)}} & \makecell[c]{\textbf{Prefill}\\\textbf{Time (ms)}} & \makecell[c]{\textbf{Avg}\\\textbf{(\%)}} \\
\midrule
FastV & 192 & 2.1 & 34.1 & 55.9 \\

PDrop & 192 & 2.0 & 36.7 & 61.6 \\

SparseVLM & 192 & 2.1 & 36.5  & 61.2 \\

\textbf{FMVR} & \textbf{36} & \textbf{0.9} & \textbf{18.0} & \textbf{62.9} \\
\bottomrule
\end{tabular}
\label{table5}
\end{table}

\begin{table*}[t]
\centering
\caption{Performance comparison on LLaVA-NeXT-7B across 10 image-based benchmarks. ‘\#VisionTokens’ is the number of vision tokens. $\Delta$ rows show the change vs.\ LLaVA-NeXT-7B baseline.}
\renewcommand{\arraystretch}{1.1}
\setlength{\tabcolsep}{3pt} 
\small
\resizebox{0.95\textwidth}{!}{
\begin{tabular}{l|c| c c c c c c c c c c | c}
\toprule
\textbf{Methods} &
\makecell[c]{\textbf{\#Vision}\\\textbf{Tokens}} &
\textbf{VQAv2} & \textbf{GQA} & \textbf{VisWiz} & \textbf{SQA\textsuperscript{IMG}} &
\textbf{VQA\textsuperscript{Text}} & \textbf{POPE} & \textbf{MME} & \textbf{MMB\textsuperscript{EN}} & \textbf{MMB\textsuperscript{CN}} & \textbf{MMVet} &
\makecell[c]{\textbf{Avg.}\\\textbf{(\%)}} \\
\midrule
LLaVA-NeXT-7B~\cite{refe30}  & 2880 & 81.3 & 62.5 & 55.2 & 67.5 & 60.3 & 86.8 & 1511.8 & 65.8 & 57.3 & 40.0 & 65.2 \\
\midrule
FastV~\cite{refe17}         &  320  & 61.5 & 49.8   & 51.3   & 66.6 & 52.2 & 49.5 & 1099.0 & 53.4   & 42.5   & 20.0   &  50.2  \\
PyramidDrop~\cite{refe18}       &  320 & 66.8 & 50.4   & 49.7   & 66.7 & 49.0 & 60.8 & 1171.5 & 55.5   & 44.7   & 24.0   & 52.6   \\
SparseVLM~\cite{refe19}        &  320 & 74.6 & 57.9 & 54.2 & 67.2 & 56.5 & 76.9 & 1386.1 & 63.1 & 56.7 & 32.8 & 60.9 \\
Prumerge+~\cite{refi22}              & 320 & 75.3 & 58.8 & 57.7 & 68.1 & 54.0 & 79.5 & 1444.3 & 63.0 & 55.6 & 31.4 & 61.6 \\
TRIM~\cite{refe20}       & 320 & 74.9 & 59.9 & 53.5 & 66.2 & 50.2 & 86.5 & 1443.8 & 63.5 & 51.0 & 32.7 & 61.1 \\
VisionZip~\cite{refe21}              & 320 & 76.2 & 58.9 & 56.2 & 67.5 & 58.8 & 82.3 & 1397.1 & 63.3 & 55.6 & 35.8 & 62.4 \\
DART~\cite{refe22}       & 320 & 75.7 & 59.5 & 56.8 & 67.5 & 57.6 & 81.0 & 1419.5 & 64.2 & 55.7 & 35.7 & 62.5 \\
DivPrune~\cite{refe23}       & 320 & 77.2 & 61.1 & 55.6 & 67.7 & 56.2 & 84.7 & 1423.3 & 63.9 & 55.7 & 34.8 & 62.8 \\
CDPruner~\cite{refe24}       & 320 & 78.4 & 61.6 & 55.8 & 67.8 & 57.4 & 87.2 & 1453.0 & 65.5 & 55.7 & 37.9 & 64.0 \\
VisPruner~\cite{refe25}       & 320 & 75.7 & 58.4 & -- & -- & 57.6 & 80.4 & 1370.1 & -- & -- & -- & -- \\
M3~\cite{refi25}        & 180 & -- & -- & -- & 72.1 & 58.7 & 87.3 & -- & 68.6 & -- & -- & -- \\
\midrule
\multicolumn{13}{l}{\textcolor{cyan!60!black}{\textbf{Ours}}}\\
 & 5 & 71.4 & 54.3 & 49.7 & 64.6 & 53.2 & 81.7 & 1365.4 & 62.3 & 52.8 & 29.4 & 58.8 \\
 & 45 & 74.1 & 57.3 & 52.4 & 66.8 & 56.4 & 84.6 & 1422.5 & 61.9 & 53.1 & 34.0 & 61.2 \\
\textbf{FMVR-LLaVA} & 180 & 78.2 & 61.4 & 55.3 & 68.6 & 58.3 & 87.6 & 1456.8 & 65.9 & 56.3 & 37.5 & \textbf{64.2} \\
 & 720 & 79.4 & 62.5 & 57.4 & 70.3 & 58.0 & 87.3 & 1487.6 & 66.5 & 57.1 & 36.9 & \textbf{65.0} \\
 & 2880 & 82.4 & 62.1 & 56.3 & 70.6 & 59.2 & 88.0 & 1526.7 & 67.1 & 56.9 & 38.2 & \textbf{65.7} \\
\bottomrule
\end{tabular}}
\label{table6}
\end{table*}

\begin{table*}[t]
\centering
\caption{Performance comparison on LLaVA-NeXT-7B across 4 video-based benchmarks. ‘\#VisionTokens’ is the number of vision tokens. $\Delta$ rows show the change vs.\ LLaVA-NeXT-7B baseline.}
\small
\setlength{\tabcolsep}{4pt}
\resizebox{0.95\textwidth}{!}{
\begin{tabular}{cccccccccccc}
\toprule
\multirow{2}{*}{\textbf{Methods}} & 
\multirow{2}{*}{\makecell[c]{\textbf{\#Vision}\\\textbf{Tokens}}} & 
\multicolumn{4}{c}{\textbf{Video-based Question-Answer}} & 
\multicolumn{6}{c}{\textbf{Video-based Generative Performance}} \\
\cmidrule(lr){3-6} \cmidrule(lr){7-12}
 &  & \multicolumn{1}{c}{\textbf{MSVD}} 
 & \multicolumn{1}{c}{\textbf{MSRVTT}} 
 & \multicolumn{1}{c}{\textbf{ActivityNet}} 
  & \textbf{Avg.}
 & \textbf{Correct.} & \textbf{Detail} & \textbf{Context.} & \textbf{Temp.} & \textbf{Consist.} & \textbf{Avg.} \\
\midrule
LLaMA Adapter~\cite{refe31} & 1280 & 54.9 & 43.8 & 34.2 & 44.3  & 2.03 & 2.32 & 2.30 & 1.98 & 2.15 & 2.16 \\
VideoChat~\cite{refe32} & 512 & 56.3 & 45.0 & 26.5 & 42.6  & 2.23 & 2.50 & 2.53 & 1.94 & 2.24 & 2.29 \\
Video-LLAMA~\cite{refe33} & 1024 & 51.6 & 29.6 & 12.4 & 31.2  & 1.96 & 2.18 & 2.16 & 1.82 & 1.79 & 1.98 \\
Video-ChatGPT~\cite{refe32} & $\sim$360 & 64.9 & 49.3 & 35.2 & 49.8  & 2.40 & 2.52 & 2.62 & 1.98 & 2.37 & 2.38 \\
BT-Adapter~\cite{refe34} & $\sim$260 & 67.5 & 57.0 & 45.7 & 56.7  & 2.68 & 2.69 & 3.27 & 2.34 & 2.46 & 2.69 \\
MovieChat~\cite{refe35} & 65536 & 75.2 & 52.7 & 45.7 & 57.9  & 2.76 & 2.93 & 3.01 & 2.24 & 2.42 & 2.67 \\
LLaMA-VID~\cite{refe36} & 1fps×2 & 69.7 & 57.7 & 47.4 & 58.3  & 2.96 & 3.00 & 3.53 & 2.46 & 2.51 & 2.89 \\
Video-LLAVA~\cite{refi14} & 2048 & 70.7 & 59.2 & 45.3 & 58.4  & 2.87 & 2.94 & 3.44 & 2.45 & 2.51 & 2.84 \\
FastV~\cite{refe17} & 198 & 38.0 & 19.3 & 30.6 & 29.3  & -- & -- & -- & -- & -- & --  \\
SparseVLM~\cite{refe19} & 198 & 68.2 & 31.0 & 42.6 & 47.3 & -- & -- & -- & -- & -- & -- \\
VisionZip~\cite{refe21} & 136 & 63.5 & 52.1 & 43.0 & 52.9  & -- & -- & -- & -- & -- & -- \\
VisPruner~\cite{refe25} & 227 & 69.3 & 55.6 & -- & --  & -- & -- & -- & -- & -- & -- \\
\midrule
\multicolumn{12}{l}{\textcolor{cyan!60!black}{\textbf{Ours}}}\\
 & 5 & 68.7 & 52.4 & 45.1 & 55.4 & 2.54 & 2.77 & 2.83 & 2.34 & 2.41 & 2.58 \\
 & 45 & 74.2 & 58.3 & 49.6 & \textbf{60.7} & 2.91 & 2.88 & 3.23 & 2.41 & 2.46 & 2.78 \\
\textbf{FMVR-LLaVA} & 180 & 76.9 & 62.5 & 51.1 & \textbf{63.5} & 3.30 & 3.55 & 3.72 & 2.84 & 3.10 & \textbf{3.30} \\
 & 720 & 78.7 & 64.8 & 53.5 & \textbf{65.7} & 3.45 & 3.62 & 3.77 & 3.05 & 3.21 & \textbf{3.42} \\
 & 2880 & 79.1 & 65.2 & 53.4 & \textbf{65.9} & 3.49 & 3.71 & 3.85 & 3.14 & 3.44 & \textbf{3.53} \\
$\Delta$ vs.\ Video-LLaVA & 180 & {\color{red}+6.2} & {\color{red}+3.3} & {\color{red}+5.8} & {\color{red}+5.1} & {\color{red}+0.43} & {\color{red}+0.61} & {\color{red}+0.28} & {\color{red}+0.39} & {\color{red}+0.59} & {\color{red}+0.46} \\
\bottomrule
\end{tabular}}
\label{table7}
\end{table*}

\begin{figure*}[t]
\centering
\includegraphics[width=\linewidth]{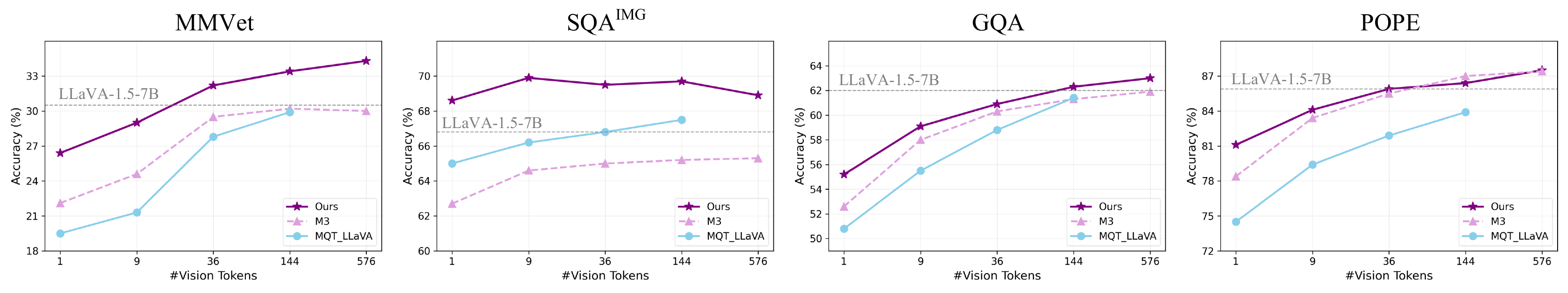}
\caption{Comparison under different numbers of vision tokens. Our method achieves higher accuracy than M3 and MQT-LLaVA.}
\label{fig4}
\end{figure*}

\begin{figure*}[t]
\centering
\includegraphics[width=0.85\linewidth]{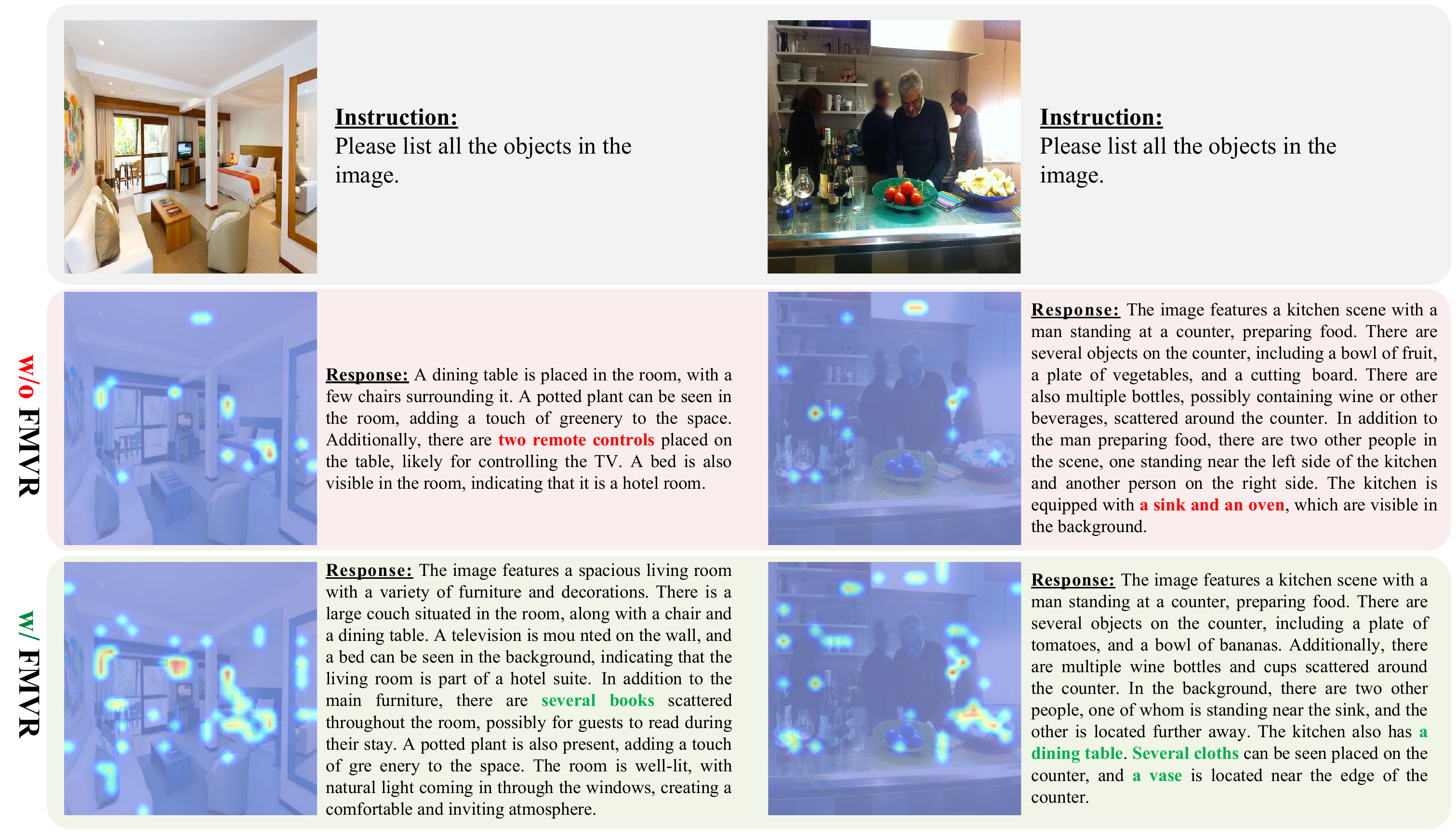}
\caption{Grad-CAM visualization and response comparison between w/o FMVR and w FMVR under 36 visual tokens. From Grad-CAM visualization, the reduction of visual tokens leads to a noticeable degradation in visual semantics, which leads to hallucinations of certain objects in the response.}
\label{fig5}
\end{figure*}

\noindent
\textbf{Implementation Details.} We implement our models based on LLaVA-1.5 \cite{refr5} and LLaVA-NeXT \cite{refi16}. They leveraging Vicuna-7B as base language model. Following \cite{refr5}, we employ a two-stage training paradigm. In the first-stage, we train only the FMVR on LLaVA-558K for one epoch with a batch size of 256 and a learning rate of 1e-3. In the second-stage, for LLaVA-1.5, we finetune FMVR and LLM using LLaVA-665K with a learning rate of LLM of 2e-5 for LLM; for LLaVA-NeXT, we finetune FMVR and LLM using LLaVA-1M \cite{refe1} with a learning rate of LLM of 1e-5. For both models, the visual encoder is optimized with a learning rate of 2e-5. Both models are trained for one epoch on 4 NVIDIA H100 GPUs.

\noindent
\textbf{Evaluation benchmarks.} We evaluate our method across 10 image-based benchmarks, including VQAv2 \cite{refe2}, GQA \cite{refe3}, VizWiz \cite{refe4}, ScienceQA-IMG \cite{refe5}, POPE \cite{refe6}, MME \cite{refe7}, MMBench \cite{refe8}, MMBench-CN \cite{refe8}, MMVet \cite{refe9}, and TextVQA \cite{refe10}. We also conduct experiments on 4 video understanding benchmarks, including MSVD-QA \cite{refe26}, MSRVTT-QA \cite{refe27}, ActivityNet-QA \cite{refe28}, and video-based generative performance benchmark \cite{refe29}.

\subsection{FMVR for Image Understanding}
Our FMVR-LLaVA achieves comparable results on 10 mainstream multimodal understanding and reasoning benchmarks. Table \ref{table1} presents the performance of different LMMs and token pruning methods. Among the average performance of 10 benchmarks, FMVR-LLaVA with only 144 visual tokens outperforms LLaVA-1.5 with 576 visual tokens by 0.8\%. Additionally, when retaining only 36 visual tokens, our method surpasses all the LLMs with vision token reduction methods in Table \ref{table1}. Notably, when drastically drop to only 1 token, our method still outperform FastV, InstructBLIP, and Qwen-VL by 1.8\%, 6.9\%, and 7.0\%, respectively. Our method also consistently outperforms MRL based methods, as shown in Fig. \ref{fig4}, under different numbers of visual tokens, our method exceeds both M3 and MQT-LLaVA on MMVet, SQA, GQA, and POPE. This means that our method can effectively improve visual semantics in the case of reduced visual tokens. 

\subsection{FMVR for high-Resolution Understanding}
Increasing the resolution of input images generally enhances the reasoning capacity of LLMs. However, this gain comes at the cost of significantly higher computational complexity. For a fair comparison, we standardize the input resolution to 672×672, corresponding to 2,880 visual tokens. Table \ref{table6} shows the performance comparison on LLaVA-NeXT-7B across 10 image-based benchmarks. Our FMVR-LLaVA (720 visual tokens) achieves comparable performance compared to LLaVA-NeXT-7B (2880 visual tokens), i.e., 65.0\% vs. 65.2\%. Even under only 180 visual tokens, FMVR-LLaVA outperforms all the visual token reduction methods, respectively. Notably, FMVR-LLaVA with only 5 visual tokens significantly exceeds FastV and PyramidDrop by 8.6\% and 6.2\%, respectively. These findings demonstrate FMVR-LLaVA’s remarkable efficiency in high-resolution multimodal reasoning scenarios.

\begin{figure*}[t]
\centering
\includegraphics[width=\linewidth]{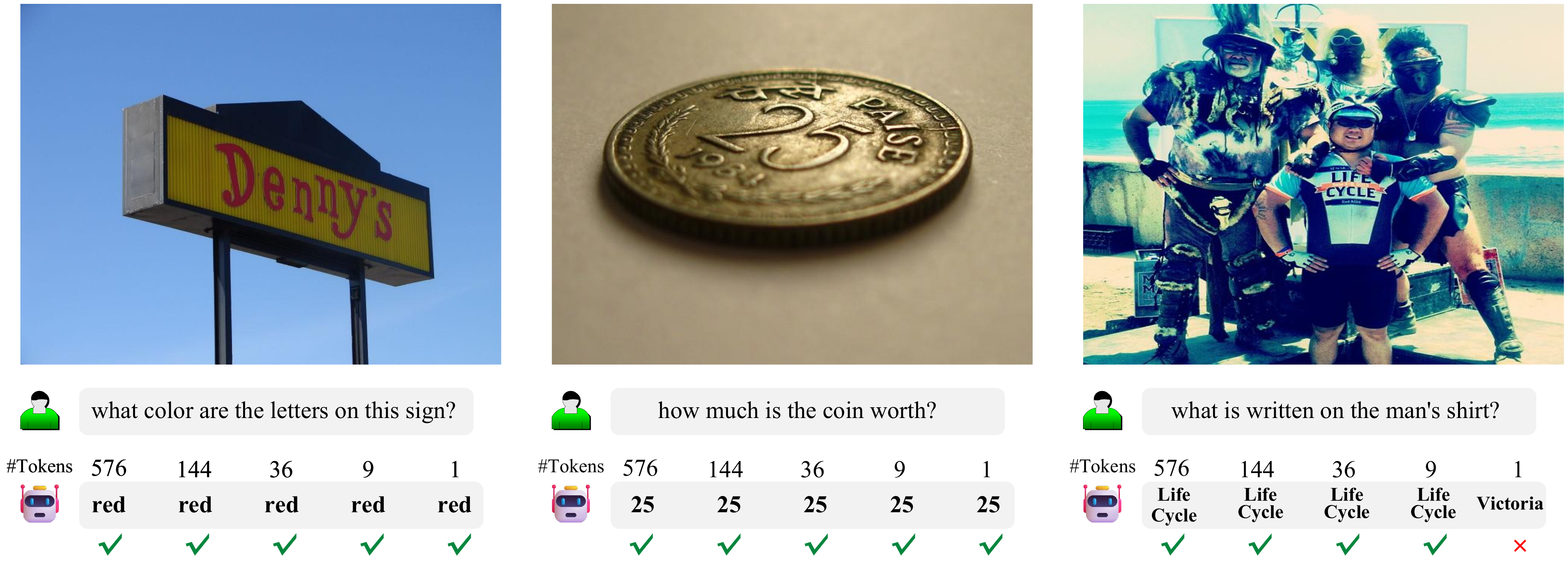}
\caption{Case of image understanding on different VQA prompts. Our method yields basically correct answers under different numbers of visual tokens.}
\label{fig6}
\end{figure*}

\subsection{FMVR for video Understanding}
We compare FMVR-LLaVA against state-of-the-art video LLMs across 4 video benchmarks. Following IG-VLM \cite{refe38}, LLaVA-NeXT-7B based FMVR-LLaVA is directly used for video reasoning. For our method, 5 frames are uniformly sampled over the entire video. As shown in Table \ref{table7}, with only 180 visual tokens, our method outperforms Video-LLaVA by 5.1\% on Video-based Question-Answer and 0.46\% on Video-based Generative Performance, respectively. These results further demonstrate the effectiveness of FMVR-LLaVA in video understanding.

\subsection{Efficiency analysis}
To demonstrate the efficiency of FMVR-LLaVA, we conduct efficiency analysis of different numbers of vision tokens on LLaVA-1.5-7B, FLOPs of each component in FMVR-LLaVA, and efficiency analysis of different pruning methods on LLaVA-1.5-7B. As shown in Table \ref{table3}, our method with 36 visual tokens achieves a ×8.9 FLOPs reduction compared to LLaVA-v1.5 while maintaining almost same performance. To further study computational load of each component, we compute the FLOPs of each module (Table \ref{table4}). The computational cost (with only 6.4e-5) of the proposed FMVR is negligible. The comparative analysis against other pruning methods showns that our method gains the best efficiency while maintaining the highest performance. Compared to FastV, our method achieves nearly a ×2.3 reduction in FLOPs and a ×1.9 reduction in Prefill time, while outperforming FastV by 7.0\%.

\subsection{Ablation Study}
We further conduct an ablation on the design of FMVR-LLaVA. For simplicity, we evaluate all the models with 144 visual tokens. As shown in Table \ref{table2}, FMVR without MaxPool unit results in performance degradation by 0.5\% due to the loss of edge feature enhancement. Similarity, the absence of AvgPool unit also lead to a 0.8\% performance degradation. In addition, compared to the baseline M3, both AvgPool and MaxPool bring performance improvements, demonstrating their importance of restoring visual semantics with reduced visual tokens.

\subsection{Case Study and Visualization Analysis}
Fig. \ref{fig6} presents examples of FMVR-LLaVA in image understanding tasks. Despite using only one visual token, our method can provide the correct answer. However, for slightly complex images, such as the third case, the answer using only one visual token is incorrect. Fortunately, increasing the number of visual tokens slightly (i.e., 9 visual tokens) can provide the correct answer. To further explore the effectiveness of FMVR, we utilize Grad-CAM \cite{refe37} to visualize the focus of visual information using 36 visual tokens. From Grad-CAM visualization results, we can see that the reduction of visual tokens leads to a noticeable degradation in visual semantics, which leads to hallucinations of certain objects in the response. On the contrary, when applying FMVR, the focus on visual semantics increases, leading to more accurate responses.

\section{Conclusion}
In this paper, we propose FMVR, a novel visual token semantic restoration method, to enhance the efficiency of LLMs under token reduction. By disentangling visual representations into low- and high-frequency components, FMVR enables the preservation of visual semantics dominated by few visual tokens and the restoration of diluted visual semantics. By integrating FMVR into MRL, it can dynamically adjust
the number of visual tokens during inference while maintaining comparable performance. Extensive experiments across 10 image and 4 video benchmarks validate the effectiveness of FMVR under token reduction.
{
    \small
    \bibliographystyle{ieeenat_fullname}
    \bibliography{main}
}
\newpage
\twocolumn[{%
\begin{center}
    {\LARGE\bfseries Frequency-Modulated Visual Restoration for Matryoshka Large Multimodal Models\par}
    \vspace{1.0em}
    {\LARGE Supplementary Material\par}
    \vspace{1.5em}
\end{center}
}]

\appendix
This supplementary material contains several sections that provide additional details related to our work. Specifically, it will cover the following topics:

\begin{itemize}
    \item In Appendix \ref{appA}, we provide details of the experimental setup, including model architectures and evaluation benchmarks. 
    \item In Appendix \ref{appB}, we provide addition experiments, including FMVR for advanced open-source LMM, FMVR for larger language model, Ablation study on other Matryoshka visual token sampling methods, and case results.
\end{itemize}

\appendix
\section{Details of Experimental Setup}
\label{appA}
\subsection{Model Architectures}
\textbf{LLaVA-1.5} \cite{refi5}. As a representative line of open-source multimodal large language models, the LLaVA family is widely adopted for its efficiency and strong capability. It mainly has three components, where CLIP \cite{refm1} is used to extract visual representations, Vicuna \cite{refs1} serves as the language backbone, and a lightweight projection module maps visual features into the LLM’s embedding space. With visual instruction tuning, this basic setup enables the model to perform a wide range of image–text tasks. LLaVA-1.5 significantly enhance multimodal reasoning ability. At the common resolution of 336×336, the model outputs 576 visual tokens for each image.

\noindent
\textbf{LLaVA-Next} \cite{refe30}. LLaVA-NeXT (also referred to as LLaVA-1.6) pushes the LLaVA architecture further by introducing a dynamic-resolution strategy aimed at strengthening the model’s visual understanding. Instead of relying on a fixed input size, it adjusts the aspect ratio according to the native dimensions of the image and allows the effective resolution to scale by up to four times. Importantly, the visual encoder itself remains unchanged. The high-resolution input is partitioned into multiple tiles, each matching the original input size, and every tile is processed independently before their visual features are concatenated and passed to the language model. This tiling-based high-resolution pipeline substantially boosts performance on tasks requiring fine-grained perception, such as OCR, detailed reasoning, and factual recognition. For consistency in our evaluations, we standardize the resolution to 672×672—four times the original—yielding a total of 2,880 visual tokens.

\noindent
\textbf{Qwen2.5-VL} \cite{refi2}. Qwen2.5-VL represents a member of Qwen-VL family. Its visual backbone has been thoroughly redesigned, adopting a revised Vision Transformer that incorporates windowed attention, SwiGLU activations, and RMSNorm, following the architectural paradigm of the Qwen2.5 language model. Beyond static image understanding, it embraces a dynamic visual processing pipeline, supporting flexible input resolutions and adaptive frame rate handling. Thanks to these advantages, Qwen2.5-VL achieves comparable performance in detailed visual perception tasks such as detection, OCR, layout and document parsing, as well as structured information extraction.

\subsection{Evaluation Benchmarks}
\subsubsection{General Image Benchmarks}
\textbf{VQAv2} \cite{refe2}. It is the version of the VQA benchmark \cite{refs2}, constructed to assess a model’s integrated understanding of visual content, natural language, and commonsense reasoning. The dataset contains 265,016 images drawn from COCO \cite{refs3} as well as synthetic abstract scenes, with each image paired with an average of 5.4 questions. Every question is further provided with 10 reference answers and 3 additional plausible distractor responses. Following common practice, we adopt the test-dev split for evaluation.

\noindent
\textbf{GQA} \cite{refe3}. It is a large-scale VQA dataset constructed from real-world images sourced from the Visual Genome corpus \cite{refe4}, aimed at evaluating the model’s capacity for compositional reasoning and fine-grained visual comprehension. It contains more than 22 million question–answer pairs, and every image is accompanied by a richly annotated scene graph that captures object categories, attributes, and inter-object relationships. For our experiments, we report results on the test-dev balanced split.

\noindent
\textbf{VizWiz} \cite{refe4}. This benchmark targets visual question answering in a real-world accessibility scenarios: blind or low-vision users take photos in everyday environments and verbally pose questions about them. Each visual question is accompanied by ten answers, offering diverse perspectives on noisy and incomplete visual information. The dataset defines two core tasks—providing answers to the visual questions and detecting when a question cannot be answered from the image alone—thereby emphasizing real-world challenges such as low-quality imagery, occlusion, and ambiguous scene content. We evaluate our model on the official test split.

\noindent
\textbf{ScienceQA} \cite{refe5}. This dataset is a large-scale multimodal multiple-choice QA benchmark that covers a broad spectrum of scientific disciplines. It comprises 21,208 questions drawn from natural sciences, language sciences, and social sciences, which are further organized into 26 topics, 127 fine-grained categories, and 379 skill types. The questions vary in modality: 48.7\% provide visual content, 48.2\% include textual context, and 30.8\% offer both forms of information. For evaluation, we follow prior works and adopt the test split that contains image-based questions (ScienceQA-IMG).

\noindent
\textbf{POPE} \cite{refe6}. This benchmark targets the evaluation of object hallucination in large vision–language models. It is constructed using images from COCO \cite{refs3}, enabling the measurement of hallucinated predictions. The performance is summarized using precision, recall, and F1 scores. Following prior works, we report results on the test split.

\noindent
\textbf{MME} \cite{refe7}. It evaluates both the perceptual and cognitive abilities of multimodal large language models across 14 subtasks. The perception tasks cover coarse- and fine-grained recognition as well as OCR, ranging from basic object presence, count, and attributes to identifying specific scenes, landmarks, and artworks. The cognition tasks assess commonsense reasoning, numerical calculation, translation, and code understanding.

\noindent
\textbf{MMBench} \cite{refe8}. This benchmark provides a broad evaluation of vision–language abilities across diverse tasks. It offers a larger varied set of questions and skills than prior benchmarks. MMBench also proposes a CircularEval protocol, which uses ChatGPT to convert open-ended outputs into structured choices for more stable and consistent scoring.

\noindent
\textbf{MM-Vet} \cite{refe9}. It examines the integration of multiple multimodal abilities. It covers six core capabilities—recognition, OCR, knowledge, language generation, spatial reasoning, and mathematics—through 218 challenging examples. Evaluation is conducted using a ChatGPT-based assessor, which provides unified metrics for answers of different formats.

\subsubsection{Text-oriented benchmarks}
\noindent
\textbf{TextVQA} \cite{refe10}. This benchmark examines the model’s capacity to recognize and reason over textual content embedded in images. The visual data, largely drawn from Open Images v3 \cite{refs5}, contains real-world settings such as signs, advertisements, and product labels, all featuring substantial scene text. As a result, the benchmark provides an assessment of OCR integration and text-sensitive visual reasoning. We evaluate on the validation split.

\noindent
\textbf{ChartQA} \cite{refs6}. It is a large benchmark for chart-based question answering, emphasizing visual understanding and logical or complex reasoning. It contains 9.6K human-written questions and 23.1K questions generated from chart summaries. Unlike earlier template-driven datasets, ChartQA requires multi-step reasoning over both chart visuals and their underlying data. We use the test split for evaluation.

\textbf{AI2D} \cite{refs7}. AI2D is a diagram-based question answering benchmark containing over 5,000 grade-school science diagrams, annotated with more than 150,000 structured labels and syntactic parses. It also provides over 15,000 multiple-choice questions paired with the diagrams, supporting research on visual reasoning and scientific diagram understanding. We use the masked test split for evaluation.

\begin{table}[t]
\centering
\caption{Performance comparison of different pruning methods on Qwen2.5-VL-7B.}
\footnotesize
\setlength{\tabcolsep}{5pt}
\renewcommand{\arraystretch}{1.2}
\resizebox{0.47\textwidth}{!}{
\begin{tabular}{l|c|cccc|c}
\toprule
\textbf{Methods} & \makecell[c]{\textbf{\#Vision}\\\textbf{Tokens}} & \textbf{TextVQA} & \textbf{ChartQA} & \textbf{AI2D} & \textbf{MMB\textsuperscript{EN}} & \makecell[c]{\textbf{Avg.}\\\textbf{(\%)}} \\
\midrule
Qwen2.5-VL  & 1296 &  84.8 & 86.1 & 80.4 & 82.8 & 83.5 \\
\midrule
FastV & 128 & 73.8 &  52.2 & 71.4 & 72.9 & 67.6 \\
DivPrune & 128 &  67.0 & 50.4 &  72.1 &  77.8 & 66.8 \\
CDPruner & 128 & 77.8 &  59.2 &  74.0 & 76.2 & 71.8 \\
\textbf{FMVR (Ours)} & 81 & 76.3 & 62.2 & 77.5 & 79.6 & \textbf{73.9} \\
\bottomrule
\end{tabular}
\label{table8}}
\end{table}

\begin{table*}[t]
\centering
\caption{Performance comparison on LLaVA-1.5-13B across 10 image-based benchmarks.  ‘\#VisionTokens’ is the number of vision tokens. $\Delta$ rows show the change vs.\ LLaVA-v1.5-13B.}
\renewcommand{\arraystretch}{1.1}
\setlength{\tabcolsep}{3pt} 
\small
\resizebox{0.95\textwidth}{!}{
\begin{tabular}{l|c| c c c c c c c c c c | c}
\toprule
\textbf{Methods} &
\makecell[c]{\textbf{\#Vision}\\\textbf{Tokens}} &
\textbf{VQAv2} & \textbf{GQA} & \textbf{VisWiz} & \textbf{SQA\textsuperscript{IMG}} &
\textbf{VQA\textsuperscript{Text}} & \textbf{POPE} & \textbf{MME} & \textbf{MMB\textsuperscript{EN}} & \textbf{MMB\textsuperscript{CN}} & \textbf{MMVet} &
\makecell[c]{\textbf{Avg.}\\\textbf{(\%)}} \\
\midrule
LLaVA-1.5-13B~\cite{refi5}         &  576  & 80.0 & 63.3   &  53.6   &   72.8 &  61.2 & 86.0 & 1531.2 & 68.5   & 63.5   & 36.2  &  66.2  \\
\midrule
FastV~\cite{refe17}         &  128  & 75.3 & 58.3   & 54.6   & 74.2 & 58.6 & 75.5 & 1460.6 & 66.1   & 62.3   & 32.8   &  63.1  \\
PyramidDrop~\cite{refe18}       &  128 & 78.2 & 61.0   & 53.8   & 73.3 & 60.2 & 83.6 & 1489.5 & 67.5   & 62.8   & 32.1   & 64.7   \\
SparseVLM~\cite{refe19}        &  128 & 77.6 & 59.6 & 51.4 & 74.3 & 59.3 & 85.0 & 1487.9 & 68.4 & 62.6 & 35.2 & 64.8 \\
Prumerge+~\cite{refi22}              & 128 & 76.2 & 58.3 & 52.8 & 73.3 & 56.1 & 82.7 & 1445.9 & 66.3 & 61.2 & 33.6 & 63.3 \\
TRIM~\cite{refe20}       & 128 & 76.4 & 59.4 & 49.7 & 72.4 & 55.0 & 86.8 & 1426.9 & 67.1 & 58.4 & 35.1 & 63.2 \\
VisionZip~\cite{refe21}              & 128 & 76.8 & 57.9 & 52.3 & 73.8 & 58.9 & 82.7 & 1449.2 & 67.4 & 62.5 & 36.0 & 64.1 \\
DART~\cite{refe22}       & 128 & 75.7 & 57.7 & 53.0 & 74.2 & 58.7 & 80.4 & 1395.0 & 65.4 & 62.2 & 34.8 & 63.2 \\
DivPrune~\cite{refe23}       & 128 & 77.1 & 59.2 & 53.5 & 72.8 & 58.0 & 86.8 & 1457.7 & 66.3 & 60.7 & 34.4 & 64.2 \\
CDPruner~\cite{refe24}       & 128 & 77.7 & 59.7 & 52.9 & 73.2 & 58.4 & 87.3 & 1478.0 & 67.5 & 61.5 & 36.2 & 64.8 \\
VisPruner~\cite{refe25}       & 128 & 76.1 & 58.8 & 53.4 & 72.8 & 57.9 & 86.0 & 1452.2 & 66.4 & 61.3 & 36.1 & 64.1 \\
MQT-LLaVA~\cite{refi26}       & 144 & 77.9 & 59.4 & 54.1 & 73.6 & 60.2 & 86.1 & 1471.3 & 66.5 & 61.0 & 31.4 & 64.4 \\
M3~\cite{refi25}              & 144 & 79.2 & 60.6 & 53.5 & 72.9 & 60.4 & 87.7 & 1478.3 & 67.2 & 59.8 & 32.3 & 64.8 \\
\midrule
\multicolumn{13}{l}{\textcolor{cyan!60!black}{\textbf{Ours}}}\\
 & 1 & 72.2 & 57.6 & 50.3 & 69.4 & 53.9 & 82.5 & 1384.7 & 62.2 & 55.8 & 27.8 & 60.1 \\
 & 9 & 76.3 & 60.2 & 51.8 & 70.9 & 58.4 & 84.7 & 1470.5 & 65.1 & 60.4 & 30.7 & 63.2 \\
\textbf{FMVR-LLaVA} & 36 & 78.4 & 61.2 & 53.4 & 71.5 & 60.5 & 86.6 & 1477.8 & 66.3 & 62.5 & 32.7 & 64.7 \\
 & 144 & 79.7 & 63.6 & 56.7 & 72.0 & 61.3 & 87.2 & 1489.4 & 68.3 & 63.8 & 34.2 & 66.1 \\
 & 576 & 80.7 & 64.2 & 57.2 & 72.4 & 60.7 & 88.1 & 1528.9 & 69.5 & 65.1 & 35.4 & \textbf{67.0} \\
\quad $\Delta$ vs.\ LLaVA-v1.5-13B & 144 & {\color{blue}-0.3} & {\color{red}+0.3} & {\color{red}+3.1} & {\color{blue}-0.8} & {\color{red}+0.1} & {\color{red}+1.2} & {\color{blue}-41.8} & {\color{blue}-0.2} & {\color{red}+0.3} & {\color{blue}-2.0} & {\color{blue}-0.1} \\
\bottomrule
\end{tabular}}
\label{table9}
\end{table*}

\begin{table*}[t]
\centering
\caption{Ablation study on Matryoshka visual token sampling, including average pooling, sequential sampling, spatial sampling, and max pooling.}
\small
\begin{tabular}{c|cccc|cccc}
\toprule
\multirow{2}{*}{Num of Vis Tokens} 
& \multicolumn{4}{c|}{TextVQA} 
& \multicolumn{4}{c}{MMBench} \\
& Avg Pooling & Sequential & Spatial & Max Pooling
& Avg Pooling & Sequential & Spatial & Max Pooling \\
\midrule
1   & 49.2 & 46.3 & 45.9 & 47.1 & 60.7 & 57.9 & 58.6 & 59.8 \\
9   & 50.8 & 46.7 & 46.4 & 48.2 & 64.2 & 60.4 & 61.5 & 63.9 \\
36  & 55.3 & 51.2 & 49.0 & 52.8 & 65.2 & 61.8 & 63.1 & 64.5 \\
144 & 55.5 & 53.7 & 50.6 & 53.3 & 65.8 & 64.2 & 63.7 & 64.9 \\
576 & 57.8 & 55.6 & 52.6 & 54.7 & 65.9 & 64.7 & 64.0 & 65.3 \\
\bottomrule
\end{tabular}
\label{table10}
\end{table*}

\begin{table*}[h]
\caption{Baseline results of token numbers on LLaVA-1.5-7B.}
\label{table11}
\centering
\small
\resizebox{0.9\textwidth}{!}{%
\begin{tabular}{c | c | c c c c c c c c c c}
\toprule
\textbf{Tokens} & \textbf{Method} & \textbf{VQAv2} & \textbf{GQA} & \textbf{VisWiz} & \textbf{SQA\textsuperscript{IMG}} & \textbf{VQA\textsuperscript{Text}} & \textbf{POPE} & \textbf{MME} & \textbf{MMB\textsuperscript{EN}} & \textbf{MMB\textsuperscript{CN}} & \textbf{MMVet} \\
\midrule
\multirow{2}{*}{576}
 & baseline     & 78.5 & 62.0 & 50.0 & 66.8 & \textbf{58.2} & 85.9 & \textbf{1510.7} & 64.3 & \textbf{58.3} & 30.5 \\
 & \textbf{Ours} & \textbf{79.2} & \textbf{63.0} & \textbf{56.5} & \textbf{68.9} & 57.8 & \textbf{87.5} & 1510.1 & \textbf{65.9} & 58.0 & \textbf{34.3} \\
\midrule
\multirow{2}{*}{144}
 & baseline    & 77.8 & 61.1 & 48.3 & 67.5 & \textbf{57.3} & 85.0 & \textbf{1480.4} & 65.1 & 57.1 & 30.2 \\
 & \textbf{Ours} & \textbf{78.6} & \textbf{62.3} & \textbf{55.1} & \textbf{69.7} & 55.5 & \textbf{86.4} & 1473.9 & \textbf{65.8} & \textbf{57.6} & \textbf{33.4} \\
\midrule
\multirow{2}{*}{36}
 & baseline    & 76.1 & 58.4 & 46.8 & \textbf{67.0} & \textbf{56.7} & 85.2 & 1448.3 & 64.5 & 57.7 & 29.4 \\
 & \textbf{Ours} & \textbf{76.5} & \textbf{60.9} & \textbf{52.9} & 69.5 & 55.3 & \textbf{85.9} & \textbf{1452.5} & \textbf{65.2} & \textbf{58.3} & \textbf{32.2} \\
\midrule
\multirow{2}{*}{9}
 & baseline    & 73.4 & 56.3 & 46.0 & 67.3 & \textbf{52.2} & 83.6 & \textbf{1422.9} & 63.7 & 56.8 & 28.5 \\
 & \textbf{Ours} & \textbf{74.5} & \textbf{59.1} & \textbf{50.7} & \textbf{69.9} & 50.8 & \textbf{84.1} & 1415.0 & \textbf{64.2} & \textbf{57.5} & \textbf{29.0} \\
\midrule
\multirow{2}{*}{1}
 & baseline    & 67.4 & 54.2 & 45.5 & 66.4 & \textbf{49.6} & 79.5 & \textbf{1291.0} & 58.2 & \textbf{53.9} & 26.2 \\
 & \textbf{Ours} & \textbf{68.3} & \textbf{55.2} & \textbf{49.7} & \textbf{68.6} & 49.2 & \textbf{81.1} & 1284.8 & \textbf{60.7} & 53.4 & \textbf{26.4} \\
\bottomrule
\end{tabular}}
\end{table*}

\begin{figure*}[t]
\centering
\includegraphics[width=0.64\linewidth]{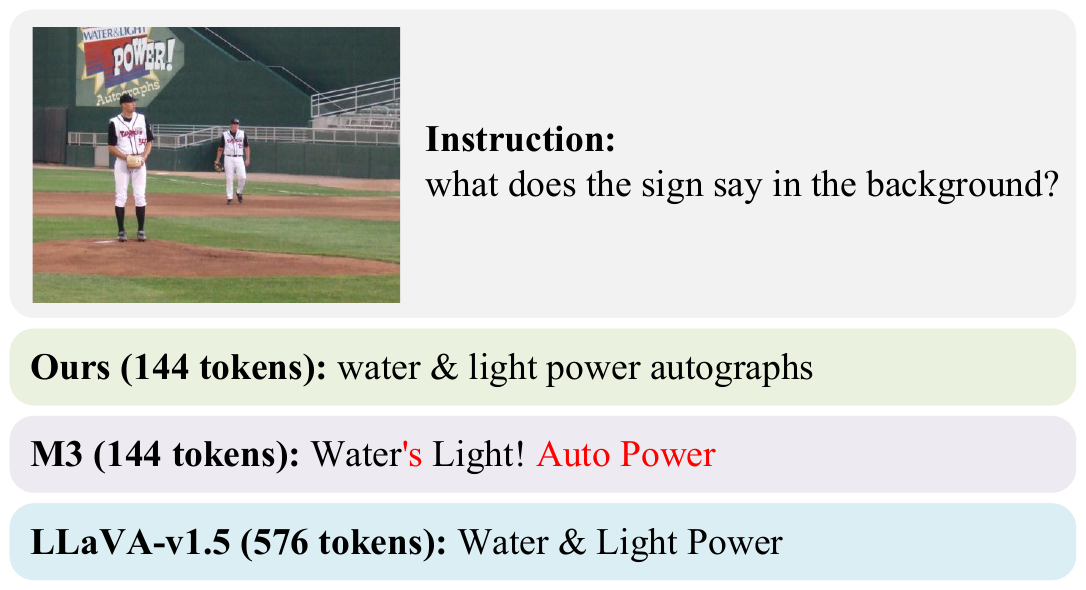}
\caption{Example demonstrating FMVR’s image understanding capability on more challenging OCR task. Response marked in red indicates errors.}
\label{fig7}
\end{figure*}

\subsubsection{Video benchmarks}
\noindent
\textbf{MSVD} \cite{refe26}. MSVD is a video question answering benchmark, containing roughly 2k short video clips and about 50k question–answer pairs. It evaluates a model’s ability to understand visual content over time, including actions, events, and temporal relationships, as well as its capacity to ground and answer natural language queries about the video.

\noindent
\textbf{MSRVTT} \cite{refe27}. It is a large-scale video captioning and video understanding dataset consisting of 10k video clips collected from real-world web videos. Each video is paired with 20 captions, covering a broad range of everyday events, activities, and scenes. The dataset is widely used for video-language tasks such as video captioning, video retrieval, and video question answering, serving as a standard benchmark for evaluating multimodal temporal understanding.

\noindent
\textbf{ActivityNet} \cite{refe28}. It is a large-scale benchmark for human activity understanding in untrimmed videos. It contains around 20k videos spanning 200 activity categories, covering a wide range of daily human actions. Designed for tasks such as action recognition, temporal action localization, and temporal segmentation, ActivityNet provides diverse and long-duration videos that evaluate both visual perception and temporal reasoning capabilities of multimodal models.

\noindent
\textbf{Video-based Generative Performance Benchmark} \cite{refe29}. It contains five dimensions: correctness, detail orientation, contextual understanding, temporal understanding, and consistency. The evaluation pipelines for both the open-ended question-answering and the generative performance benchmarks adhere to Video ChatGPT \cite{refe29}.

\section{Additional Experimental Results}
\label{appB}

\subsection{FMVR for advanced open-source LMM}
In addition to LLaVA, we further apply FMVR to one of the most advanced open-source LMM, i.e., Qwen2.5-VL. The input image is resized to 1008×1008 and the visual encoder produces a total of 1,296 tokens. We directly insert the trained FMVR into the back of the Qwen2.5-VL's visual encoder. We compare our FMVR with FastV DivPrune, and CDPruner. As shown in Table \ref{table8}, FMVR consistently achieves the best results across on four benchmarks. FMVR with 81 visual tokens outperforms the next-best method, CDPruner with 128 visual tokens, by 2.1\% in average. It demonstrates the strong generalizability of FMVR on advanced LLM architectures.

\subsection{FMVR for larger language model}
To further examine whether our approach scales to stronger language backbones, we conduct experiments on LLaVA-1.5-13B. As shown in Table \ref{table9}, Increasing the size of LLM yields substantial performance gains. Across all token pruning methods with 128 visual tokens except for SparseVLM, CDPruner, and M3, FMVR with only 36 visual tokens consistently achieves the highest performance 64.7\%. It is worth noting that our FMVR with 144 visual tokens achieved comparable performance to LLaVA-1.5-13B with 576 visual tokens, i.e., 66.1\% vs. 66.2\%, demonstrating its effectiveness on larger language models.

\subsection{Ablation study on other Matryoshka visual token sampling methods}
We evaluate four strategies for selecting visual tokens in Matryoshka visual token construction, including average pooling, spatial sampling, sequential sampling, and max pooling. As illustrated in Table \ref{table10}, average pooling consistently outperforms the other three approaches because it preserves the complete local semantic information by aggregating all tokens within each spatial region. In contrast, spatial sampling and sequential sampling select only one token from each region or sequence, which discards important information and disrupts spatial consistency. Max pooling keeps only the strongest activation, ignoring other semantic information. As a result, average pooling provides the most informative representation, leading to superior performance.

\subsection{Additional case results}
Fig. \ref{fig7} presents an example that highlights FMVR's strength on a challenging OCR task. The text in the image varies in size and is somewhat obscured, which typically degrades recognition performance in LLMs. Despite this difficulty, our FMVR correctly read all the text using only 144 visual tokens. In contrast, M3 misidentify key words, such as mistakenly identifying 'autographs' as'auto power' and introduce an incorrect ''s'. Furthermore, LLaVA-v1.5 fails to recognize 'autographs'. These results demonstrate that FMVR possesses more detailed visual perception capabilities.

\subsection{Baseline results of token numbers}
Table \ref{table11} reports the baseline and our method under different token budgets on LLaVA-1.5-7B. As token numbers decrease from 576 to 1, performance gradually drops. However, our method consistently outperforms the baseline on most benchmarks, demonstrating stronger robustness and effectiveness under aggressive token reduction.

\end{document}